%% file: main.tex
\documentclass[runningheads]{llncs}

\usepackage[mobile]{eccv}

\usepackage{eccvabbrv}
\usepackage{graphicx}
\usepackage{booktabs}
\usepackage{multirow}
\usepackage{pifont}
\usepackage{wrapfig}
\usepackage[bottom]{footmisc}
\usepackage[accsupp]{axessibility}  

\usepackage[pagebackref,breaklinks,colorlinks,citecolor=eccvblue]{hyperref}

\usepackage{orcidlink}
\usepackage{soul}

\input{preamble}

\begin{document}

\title{\texttt{CrossScore}: Towards Multi-View Image Evaluation and Scoring}


\author{
Zirui Wang\quad
Wenjing Bian\quad
Victor Adrian Prisacariu
}

\authorrunning{Z.~Wang et al.}

\institute{
    Univeristy of Oxford\\
    \email{\{ryan, wenjing, victor\}@robots.ox.ac.uk}\\
    \url{https://crossscore.active.vision}
}

\maketitle

\input{00_abs}
\input{01_intro}

\input{02_related}
\input{03_method}

\input{04_exp}

\input{05_conclusion}
\input{06_ack}
\input{08_supp}
\input{07_reference}

\end{document}

%% file: preamble.tex
\usepackage[dvipsnames]{xcolor}

\newcommand*{\ourscore}{{\texttt{CrossScore}}\@\xspace}

\newcommand{\cmark}{\ding{51}}
\newcommand{\xmark}{\ding{55}}


%% file: 00_abs.tex
\input{figures/00_teaser}

\begin{abstract}
    We introduce a novel \textit{cross-reference} image quality assessment method that effectively fills the gap in the image assessment landscape, complementing the array of established evaluation schemes -- ranging from
    \textit{full-reference} metrics like SSIM~\cite{wang2004image}, 
    \textit{no-reference} metrics such as NIQE~\cite{mittal2012making}, to 
    \textit{general-reference} metrics including FID~\cite{heusel2017gans}, and 
    \textit{Multi-modal-reference} metrics, \eg CLIPScore~\cite{hessel2021clipscore}. 
    Utilising a neural network with the cross-attention mechanism and a unique data collection pipeline from NVS optimisation, our method enables accurate image quality assessment without requiring ground truth references.
    By comparing a query image against multiple views of the same scene, our method addresses the limitations of existing metrics in novel view synthesis (NVS) and similar tasks where direct reference images are unavailable.
    Experimental results show that our method is closely correlated to the full-reference metric SSIM, while not requiring ground truth references.
\end{abstract}

%% file: figures/00_teaser.tex
\vspace{-0.7cm}
\begin{figure}[h]
    \centering
    \includegraphics[width=0.8\columnwidth]{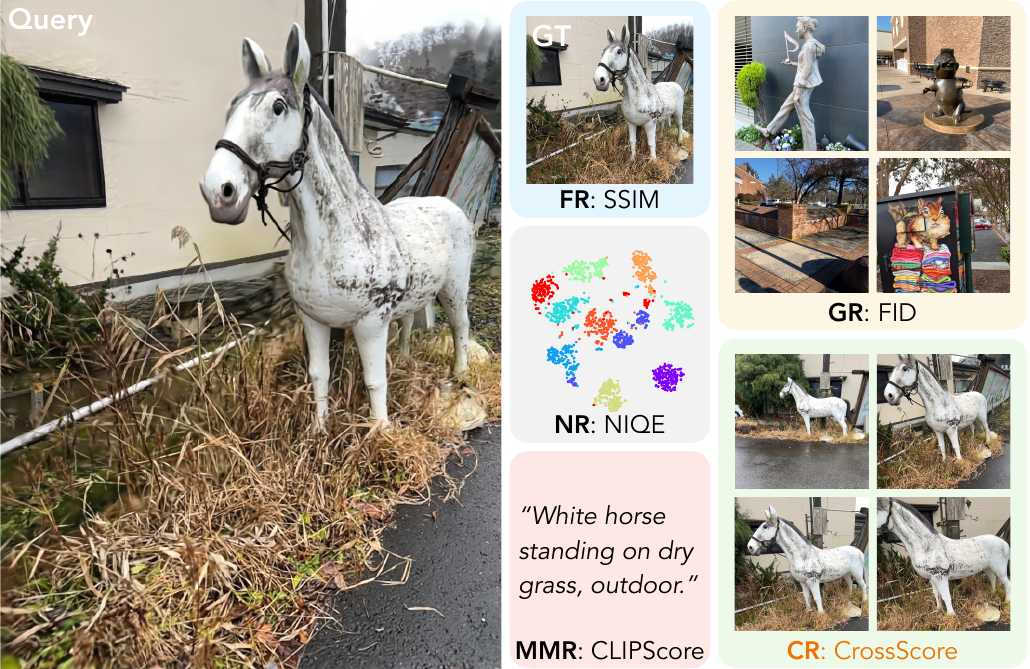}
    \caption{
        We propose a novel \textit{cross-reference} (\textcolor{Orange}{CR}) image quality assessment (IQA) scheme, which evaluates a query image using multiple unregistered reference images that are captured from different viewpoints. 
        This approach sets a new research trajectory apart from conventional IQA schemes such as \textit{full-reference} (FR), \textit{general-reference} (GR), \textit{no-reference} (NR), and \textit{multi-modal-reference} (MMR).
    }
    \label{fig:teaser}
\end{figure}
\vspace{-1cm}

%% file: 01_intro.tex
\section{Introduction}
\label{sec:intro}

Accurate image quality evaluation is critical for enhancing the performance of computer vision tasks, including image processing, image generation, and novel view synthesis (NVS).

Image quality assessment (IQA) methods can be categorised based on the type of referencing.
The most popular group, \textit{full-reference} (FR) metric, such as PSNR, SSIM~\cite{wang2004image}, LPIPS~\cite{zhang2018unreasonable}, evaluates the differences between a query image and a reference image in terms of pixels and perceptual quality. These metrics are essential for tasks such as super-resolution, denoising, and compression, and, importantly, assume the availability of an oracle reference image.

Ground truth images, however, may not always be available, for instance, in image generation tasks. 
Consequently, multiple attempts have been made to alleviate the dependency on ground truth oracles. 
For instance, FID~\cite{heusel2017gans} is a \textit{general-reference} (GR) scheme that evaluates the discrepancy of data distribution between image sets. 
Alternatively, \textit{multi-modal-reference} (MMR) approaches, such as CLIPScore~\cite{hessel2021clipscore}, examine the image-text similarity, whereas \textit{no-reference} (NR) metrics, such as NIQE~\cite{mittal2012making} and PIQE~\cite{venkatanath2015blind} evaluate single-image statistics without referencing. 

Although the aforementioned non-ground-truth metrics are widely employed in tasks like image compression, denoising, and generation, these methods generally rely on high-level statistics and global context, consequently lacking the capacity for detailed analysis. This deficiency renders them inadequate for NVS, which requires pixel-level assessment of novel view images with scene-specific context.

The established approach to assess NVS performance involves selecting a subset of test images from an existing camera trajectory, which cannot be used in training, rendering images using their camera parameters, and computing pixel-level FR scores by comparing these rendered images with the original captured test images.
While generally simple and effective, this subsampling approach exhibits two primary issues:
1) balancing the number of images between training and evaluation can affect the statistical relevance of the assessment and the effectiveness of the training; and
2) relying on FR metrics precludes the ability to evaluate renderings using true novel trajectories, as ground truth images are not available for true novel views.

\ul{These challenges motivate us to develop a novel IQA scheme, which evaluates the quality of a query image using multiple reference views, each observing the same content but from different viewpoints}. 
The key intuition is to leverage multi-view images as a substitute for a ground truth image, enabling a `perspective'-version FR evaluation.
We term this process as \textit{cross-reference} (CR) evaluation and the resulting score as \texttt{CrossScore}.
Our method differs from prior works in two essential ways: 
1) unlike FR-IQA, our approach eliminates the need for aligned reference images; and
2) in contrast to GR-, NR-, and MMR-IQA, our method offers detailed evaluation via multi-view reasoning.

Specifically, we propose to find a cross-reference function that predicts a full-reference metric, \ie SSIM, of a distorted image, by comparing it with multiple unregistered reference images.
We implement this function with a neural network that employs the cross-attention mechanism~\cite{vaswani2017attention}. 

Alongside the model formulation, a key additional challenge lies in gathering training samples.
Our solution involves rendering images throughout the NVS optimisation process, covering a broad spectrum of distortion varieties and intensities. 
By comparing these images against their original, undistorted versions, we obtain pixel-level SSIM scores that serve as a training objective. 
This self-supervised data collection approach allows us to build a rich dataset and enables per-pixel supervision for our network training.

In summary, our contribution is threefold.
\textit{First}, we unveil CR-IQA, a new image evaluation regime tailored for multi-view scenarios.
\textit{Second}, we actualise this concept through a neural network based on cross-attention mechanisms, enabling detailed per-pixel evaluation in the absence of ground truth images.
\textit{Third}, we develop a self-supervised data collection scheme that utilises existing NVS algorithms to produce a wide variety of distorted images along with their SSIM score maps, serving as our training samples.
Our findings demonstrate that the \ourscore aligns closely with the full-reference SSIM score, while eliminating the need for ground truth reference images.

%% file: 02_related.tex
\section{Related Work}
\label{sec:related_work}

This section offers an overview of image quality assessment (IQA) metrics, sorted by reference image availability and nature, and reviews the current evaluation framework for novel view synthesis (NVS).

\subsection{Image Quality Assessment Metrics}
\subsubsection{Full-Reference Metrics}
Mean Squared Error (MSE), Peak Signal-to-Noise Ratio (PSNR), and Structural Similarity Index Measure (SSIM)~\cite{wang2004image} are key metrics in the FR-IQA group for comparing a query image with the ground truth due to their simplicity and accuracy. 
Further, several variants are proposed to improve performance by evaluating at multi-scale~\cite{wang2003multiscale}, utilising handcrafted features~\cite{xue2013gradient,zhang2011fsim} and extend to specific applications, such as image stitching~\cite{solh2009miqm}, and high dynamic range (HDR) images~\cite{mantiuk2011hdr}.
Recently, deep neural networks have advanced FR-IQA towards aligning assessments more closely with human visual perception~\cite{zhang2018unreasonable, ding2020image}. 
Overall, FR-IQA offers detailed evaluation at the cost of requiring ground truth images.

\subsubsection{Reduced-Reference Metrics}
RR-IQA methods are designed to address situations where only partial information about the original reference image is accessible. 
Wang \etal~\cite{wang2005reduced, wang2006quality} uses generalised Gaussian density model parameters that model natural images in the wavelet transform domain as RR features. 
Reduced-reference structural similarity index (RR-SSIM)~\cite{rehman2012reduced} approximates FR-SSIM by using image statistical properties in the divisive normalisation transform domain. 
Redi \etal~\cite{redi2010color} uses descriptors based on colour correlogram that describes the spatial correlation of colours as RR data. 
RR metrics are commonly used for reducing transmission costs or accelerating processing, with a trade-off in information. Since the reference features still come from the ground truth image, the RR group shares the same limitation as FR-IQA in many applications.

\subsubsection{No-Reference Metrics}
NR-IQA methods provide an alternative to evaluating image quality based on the input only when ground truth references are unavailable. 
Classical approaches like DIVINE~\cite{moorthy2011blind}, BRISQUE~\cite{mittal2012no}, NIQE~\cite{mittal2012making} PIQE~\cite{venkatanath2015blind} are designed with handcrafted features and natural scene statistic models to capture image distortions and estimate quality.
Kang \etal~\cite{kang2014convolutional} first applied CNN to NR-IQA. TRIQ~\cite{you2021transformer} applied a transformer encoder to features extracted by CNN to predict image quality. 
MUSIQ~\cite{ke2021musiq} addressed the CNN size constraint with a patch-based transformer. 
NR metrics are primarily tailored to measure distortions, including compression artefacts, noise, and blur. 
However, their ability to provide a comprehensive analysis of image content is limited by the lack of reference images, rendering them less suitable for multi-view scenarios.

\subsubsection{General-Reference Metrics}
To assess the quality of generated images, commonly used metrics include Inception Score (IS)~\cite{barratt2018note, salimans2016improved}, FID~\cite{heusel2017gans} and KID~\cite{binkowski2018demystifying, xu2018empirical}. 
These metrics evaluate the overall performance of generative models rather than scoring individual images.
For instance, FID measures the
squared Fr\'echet distance between the distributions of the reference image set and the generated image set. 
These metrics focus on global statistics, making them well-suited for assessing image generation models~\cite{goodfellow2014generative, ho2020denoising} while infeasible for novel view synthesis tasks~\cite{mildenhall2021nerf, mildenhall2019local}.

\subsubsection{Multi-Modal-Reference Metrics}
Cross-modal models facilitate the assessment of alignment between images and text, enabling the development of an MMR-IQA scheme.
CLIPScore~\cite{hessel2021clipscore} directly applies the CLIP model~\cite{radford2021learning} to the image captioning task by computing the adjusted cosine similarity between the image and candidate text as a score. 
LIQE~\cite{zhang2023blind} employs a multi-task learning model leveraging vision-language correspondences to estimate the quality score, scene category and distortion type.
CLIP-IQA~\cite{wang2023exploring} applies CLIP to IQA with simple antonym prompts to access image qualities such as brightness and noisiness.
By associating semantics between text and vision, these metrics are commonly used in text-to-image generation and editing~\cite{kawar2023imagic, saharia2022photorealistic} and image captioning tasks~\cite{nguyen2024improving}, yet they lack the capability for detailed evaluation.

\subsection{Image Quality Assessment in NVS Systems}
Common evaluation metrics for NVS tasks include Full-Reference (FR) metrics such as PSNR, SSIM~\cite{wang2004image}, and LPIPS~\cite{zhang2018unreasonable}, which produce detailed similarity assessment between rendered and ground truth images. 
Enhancements to the evaluation process have been proposed, including introducing explicit representation~\cite{azzarelli2023towards}, simplifying evaluation through metric summarisation~\cite{barron2021mip}, incorporating additional robustness metrics~\cite{wang2023benchmarking}, and benchmarking with more effective camera coverage~\cite{de2023scannerf}.

As NVS rapidly evolves to address more complex tasks, conventional FR-style evaluations struggle, particularly with novel views lacking ground truth camera data, as seen in tasks like joint camera parameter and NeRF optimisation~\cite{wang2021nerfmm,lin2021barf,jeong2021self,truong2023sparf,park2023camp,bian2023porf,chen2023dbarf}. Additionally, large-scale scenes and dramatic camera movements, such as in city-scale~\cite{tancik2022block,xiangli2022bungeenerf,rematas2022urban} and egocentric setups~\cite{pan2023aria, sun2023aria, somasundaram2023project, tschernezki2021neuraldiff,deng2022fov, grauman2022ego4d, Damen2021epic}, render the subsample-then-compare strategy inadequate. These issues underscore the need for a metric better suited to multi-view evaluation while ground truth is not available. 

%% file: 03_method.tex
\input{figures/01_method}
\section{Method}
\label{sec:method}

Our goal is to evaluate the quality of a query image $\Tilde{I}_q$, using a set of reference images $\mathcal{I}_{r} = \{I_{r}^{i} | i=1...N_{\text{ref}}\}$ that capture the same scene as the query image but from other viewpoints, where $i$ denotes the $i^{th}$ image in a reference set with $N_{\text{ref}}$ images.
We refer to this reference set as the \textit{cross-reference} (CR) set.
From the NVS application perspective, the query image $\Tilde{I}_q$ is often a rendered image with artefacts, and the reference set consists of the real captured images.

To achieve this goal, we propose a simple but effective strategy, by finding a function that predicts a well-established FR score, \eg SSIM, for a query image.
Unlike the SSIM function, which takes the input of a pre-aligned ground truth image, our new function takes multi-view images as input.

For a query image $\Tilde{I}_q \in \mathbb{R}^{H \times W \times 3}$, the SSIM function compares it with its ground truth image $I_q$ in a sliding window fashion, and outputs a score map 
$\mathbf{S}_{\text{ssim}} \in \mathbb{R}^{H \times W}$:
\begin{equation}
    f(\Tilde{I}_q, I_q) \mapsto \mathbf{S}_{\text{ssim}},
\end{equation}
where $f(\cdot)$ denotes the SSIM function.\footnote{
The raw SSIM score map shares the same dimensions as a query image, \ie $\mathbf{S}_{\text{ssim}} \in \mathbb{R}^{H \times W \times 3}$. For simplicity, we follow a standard practice that averages the SSIM scores across colour channels, yielding a single-channel score map $\mathbf{S}_{\text{ssim}} \in \mathbb{R}^{H \times W}$.
}

The aim of our work is to predict a score map $\mathbf{S}_{\text{cross}} \in \mathbb{R}^{H \times W}$, which is highly correlated with $\mathbf{S}_{\text{ssim}}$, by comparing a query image with the CR set $\mathcal{I}_{r}$, instead of with its fully aligned ground truth image $I_q$. 
In other words, we seek a function $g(\cdot)$ that approximates the SSIM function $f(\cdot)$, but making use of the CR set $\mathcal{I}_{r}$:
\begin{equation}
    g(\Tilde{I}_q, \mathcal{I}_{r}) \mapsto \mathbf{S}_{\text{cross}} \approx \mathbf{S}_{\text{ssim}}.
\end{equation}
The intuition here is to approximate a `perspective SSIM' function by replacing the ground image with a set of unregistered multi-view images.

We parameterise the cross-reference function $g(\cdot)$ with a neural network $\Phi$. 
We elaborate on the network design and training strategy in the subsequent sections.

\subsection{Network Design}
\label{sec:method:network}
As shown in \cref{fig:method}, our network $\Phi$ consists of three parts, 
i) an image encoder $\Phi_{\text{enc}}$, which extracts feature maps from input images; 
ii) a cross-reference module $\Phi_{\text{cross}}$, which associates a query image $\Tilde{I}_q$ with images in a CR set $\mathcal{I}_{\text{ref}}$ and produces a latent score map; and 
iii) a score regression head $\Phi_{\text{dec}}$ that decodes the latent score map to the final score map $\mathbf{S}_{\text{cross}}$.

\subsubsection{Image Encoder $\Phi_{\text{enc}}$}
We adapt a pre-trained DINOv2~\cite{oquab2023dinov2} network as our image encoder $\Phi_{\text{enc}}$, which takes an image $I$ as input and outputs a feature map $\mathbf{F} = \Phi_{\text{enc}}(I)$. 
This image encoder is applied to all images including query and reference images, and produces feature maps $\mathbf{F}_q$ and $\mathbf{F}_r^i$.
We adopt the same patch-wise positional encoding scheme as DINOv2, with each small patch being assigned a positional embedding. 
Since our cross-reference function takes a set of unordered reference images, image-wise encoding is not applied.

\subsubsection{Cross-Reference Module $\Phi_{\text{cross}}$}
We leverage a Transformer Decoder~\cite{vaswani2017attention} in our cross-reference module $\Phi_{\text{cross}}$.
Given a feature map of a query image $\mathbf{F}_q$, the cross-reference module $\Phi_{\text{cross}}$ outputs a latent score map $\mathbf{M} = \Phi_{\text{cross}}(\mathbf{F}_q, \mathcal{F}_r)$, by comparing $\mathbf{F}_q$ with the set of reference feature maps $\mathcal{F}_r = \{ \mathbf{F}_r^i | i=1...N_{\text{ref}} \}$.
Specifically, this cross-reference is conducted by the cross-attention mechanism, where the feature map of query image $\mathbf{F}_q$ is the \textit{query} of the cross-attention, and the set of feature maps of the reference images $\mathcal{F}_r$ serve as \textit{key} and \textit{value} in the cross attention. 

\subsubsection{Score Regression Head $\Phi_{\text{dec}}$}
With a latent score map predicted from the cross-reference module $\Phi_{\text{cross}}$, a small regression head $\Phi_{\text{dec}}$ is applied to finally predict \texttt{CrossScore} $\mathbf{S}_{\text{cross}}$.
We use a shallow Multi-layer Perceptron (MLP) to interpret a latent score map to a per-pixel score map.
Since DINOv2 encodes images by patches, a latent score vector contains the quality estimation for the entire patch. 
In order to predict \ourscore in pixel level, we use the last layer of the MLP to interpret each latent score estimation to a 196-dimension vector, which is then reshaped to a $14 \times 14$ patch that corresponds to a small image patch encoded by DINOv2.

\subsection{Training Strategy}
\label{sec:method:training_strategy}
\subsubsection{Self-supervised Training Data Collection}
We leverage existing NVS systems and abundant multi-view datasets to generate SSIM maps for our training.
Specifically, we select Neural Radiance Field (NeRF)-style NVS systems as our data engine.
Given a set of images, a NeRF recovers a neural representation of a scene by iteratively reconstructing the given image set with photometric losses.
By rendering images with the camera parameters from the original captured image set at multiple training checkpoints, we generate a large number of images that contain various types of artefacts at various levels. 
From which, we compute SSIM maps $\mathbf{S}_{\text{ssim}}$ between rendered images and corresponding real captured images, which serve as our training objectives.

\subsubsection{Supervision}
This data collection scheme enables a self-supervised training scheme 
for our network $\Phi$. We consider each rendered image as a query image $\Tilde{I}_q$, and we randomly sample real captured images (exclude the real image of the query) to form our reference set $\mathcal{I}_{r}$.
The SSIM map of the rendered image $\mathbf{S}_{\text{ssim}}$ is then used to supervise our network to predict a $\mathbf{S}_{\text{cross}}$ 
with an $L_1$ loss:
\begin{equation}
    \mathcal{L} = | \mathbf{S}_{\text{ssim}} - \mathbf{S}_{\text{cross}} |.
\end{equation}

%% file: figures/01_method.tex
\begin{figure}[t]
    \centering
    \includegraphics[width=1.0\columnwidth]{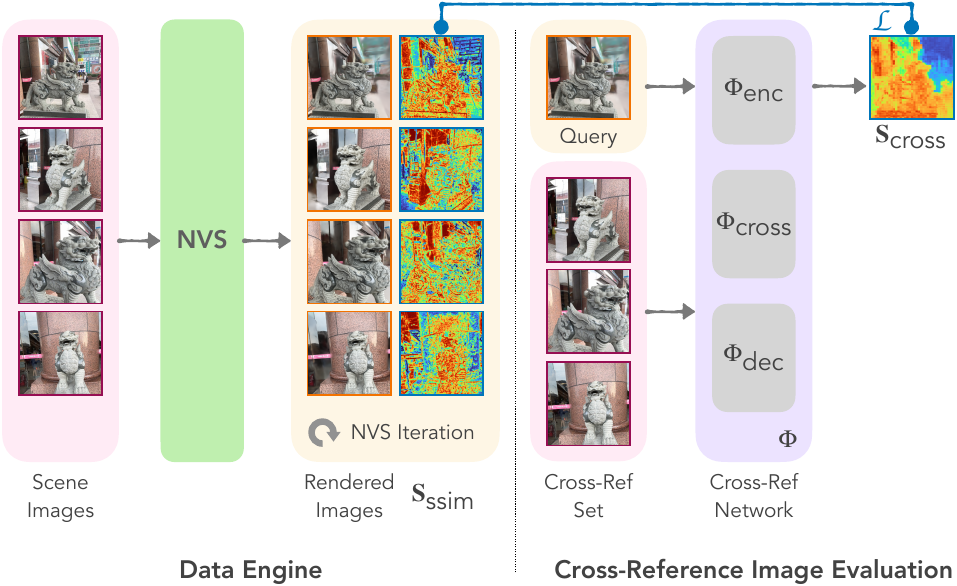}
    \caption{
    \textbf{Data Generation and Training Pipeline.} 
    We employ existing NVS models to generate pairs of rendered images and SSIM maps for training purposes. As the NVS model iterates, rendered images at various optimisation stages are used as the query image for input into our model. Together with a set of reference images from the same scene, our model predicts a score map, supervised by the corresponding SSIM map.
    More details see \cref{sec:method:network,sec:method:training_strategy}.
    }
    \label{fig:method}
\end{figure}

%% file: 04_exp.tex
\section{Experiments}
\label{sec:experiments}
We start this section by outlining our experimental setup in \cref{sec:exp:setup}, followed by assessing the correlation between our score and SSIM through both qualitative and quantitative analyses in \cref{sec:exp:main_corr}. 
We then demonstrate the application of our CR-IQA in two scenarios: 
benchmarking unseen NeRF algorithms (\cref{sec:exp:app:few_shot_nerf}) and
evaluating images rendered from novel trajectories in NVS without ground truth (\cref{sec:exp:app:novel_traj}).
Additionally, we examine the effectiveness of our cross-reference module via a visualisation of its attention maps in~\cref{sec:exp:vis_attn} and an ablation study in \cref{sec:exp:ablation}. Lastly, \cref{sec:exp:limitation_and_future} concludes our experiments with a discussion on limitations and future research directions.

\subsection{Experimental Setup}
\label{sec:exp:setup}

\subsubsection{Dataset}
We utilise three datasets in our primary experiments. 
First, the Map-free Relocalisation (\ul{MFR})~\cite{arnold2022mapfree} dataset, initially designed for camera parameter estimation benchmarks, has been adapted for our data collection and network training. This dataset features 460 outdoor videos of objects and buildings in a resolution of $540 \times 960$. 
Second, we utilise the \ul{Mip360}~\cite{barron2022mipnerf360} dataset, consisting of 9 videos that capture 360-degree scans of diverse scenes, both outdoor and indoor.
The original resolution of the images is $\sim$4K. To facilitate DINOv2~\cite{oquab2023dinov2} image encoding, we downscale all images by a factor of 4.
Third, we randomly select 10 videos from the RealEstate10K (\ul{RE10K})~\cite{zhou2018stereo} dataset, originally in $1920\times1080$ resolution, which are downscaled to $960 \times 540$. 
\ul{Training and evaluation}:
Our network is solely trained on the MFR dataset, from which 348 and 14 videos are randomly selected as training and evaluation split respectively.
In addition to evaluating on MFR evaluation split, we further assess the performance of our method using Mip360 and RE10K datasets.

\subsubsection{Metrics}
To evaluate the effectiveness of our method in predicting scores closely aligned with SSIM values, we use correlation coefficients as our primary evaluative metric.
The Pearson correlation coefficient~\cite{pearson1896vii} is utilised across the majority of our analyses, complemented by Spearman's rank correlation coefficient~\cite{spearman1961proof} for studies involving new camera trajectories.  
These coefficients, ranging in $[-1, 1]$, measure the strength of association between \ourscore and SSIM, with a larger magnitude indicating a stronger correlation.

\subsubsection{Baselines}
We choose five well-established IQA methods as baselines. Two from FR-IQA family: SSIM~\cite{wang2004image} and PSNR, and three from NR-IQA family: BRISQUE~\cite{mittal2012no}, NIQE~\cite{mittal2012making}, and PIQE~\cite{venkatanath2015blind}. 

\subsubsection{Network Training}
\ul{Architecture}:
We adopt a pre-trained DINOv2-small network as the image encoder $\Phi_{\text{enc}}$, which encodes images with a patch size $14\times14$ and produces features in 384 channels.
The $\texttt{CLS}$ token is ignored.
The cross-reference module $\Phi_{\text{cross}}$ incorporates 2 transformer decoder layers with hidden dimension 384, and the decoder $\Phi_{\text{dec}}$ is equipped with a 2-layer MLP. 
\ul{Pre-processing}:
During training, we randomly crop $518\times518$ images from both query and reference images, whilst during inference, our model supports inputs at arbitrary resolutions. 
Notably, raw SSIM maps may occasionally present values below zero, and we found clamping raw SSIM maps to the range $[0, 1]$ leads to a slightly more stable training process.
For the cross-reference set selection, we randomly choose $N_\text{ref} = 5$ real images from the same scene as the query image.
\ul{Optimisation}:
We apply a constant learning rate of 5e-4 with an Adam-W~\cite{loshchilov2018decoupled} optimiser, training on $2\times$ NVIDIA A5000 24GB GPUs for 160,000 iterations in 60 hours, with a per-GPU batch size of 24.

\subsubsection{Training Data Generation}
We optimise Gaussian-Splatting (GS)~\cite{kerbl3Dgaussians}, Nerfacto~\cite{tancik2023nerfstudio}, and TensoRF~\cite{Chen2022ECCV} on the MFR~\cite{arnold2022mapfree} dataset for 15,000 iterations, saving checkpoints every 1,000 iterations up to 10,000, and a final one at 15,000 iterations. Images rendered from these checkpoints are compared against ground truths to produce SSIM maps. To reduce the cost of this process, we temporally subsample the MFR dataset by a factor of 8. The entire data processing spanned approximately two weeks, utilising $4\times$ NVIDIA A5000 GPUs. The generated images and SSIM maps take about 1.5TB of storage.
Our selection of GS, Nerfacto, and TensoRF was based on their efficiency and output quality. 
Each method employs a distinct NVS approach: GS models scenes with point clouds, Nerfacto utilises voxel grids, and TensoRF decomposes a 3D scene to planes, ensuring a diverse and high-quality image rendering process. 
As a result, this approach balances data generation cost while producing a wide variety of distorted images and accurate corresponding SSIM maps.

\subsection{Correlation with SSIM}
\input{figures/04_main_results}
\input{tables/main_correlation}

\label{sec:exp:main_corr}
We evaluate \ourscore by comparing it to SSIM using Pearson Correlation~\cite{pearson1896vii}, with results shown in \cref{tab:main_correlation} alongside other baselines. Our approach demonstrates a strong correlation with the full-reference SSIM without using ground truth. 
Moreover, trained solely on the MFR dataset, our method successfully generalises to various settings, including indoor, outdoor, and 360-degree scanning environments, highlighting its versatile applicability. 
\cref{fig:exp:main_results} provides qualitative results supporting our findings.

\subsection{Application: Evaluating Few-shot NeRFs}
\label{sec:exp:app:few_shot_nerf}
This experiment demonstrates the application of \ourscore for evaluating few-shot NeRF methods, specifically comparing IBRNet~\cite{wang2021ibrnet} and PixelNeRF~\cite{yu2021pixelnerf} using official checkpoints. \cref{tab:exp:ibrnet_vs_pixelnerf} shows that both SSIM, PSNR, and \ourscore suggest IBRNet performs better on the MFR dataset. Note that the aim here is to highlight the ability of \ourscore to discern performance differences between methods rather than to benchmark them comprehensively. 
\input{tables/ibrnet_vs_pixelnerf}

\subsection{Application: IQA on Images Rendered From a Novel Trajectory}
\label{sec:exp:app:novel_traj}
\input{figures/06_novel_traj}
\input{tables/novel_trajectory}
In this experiment, we demonstrate that our cross-reference method enables true novel view rendering evaluation. 
Specifically, given an NVS-reconstructed scene, we evaluated this scene in two distinct ways, as illustrated in \cref{fig:exp:novel_traj_setup}.
First, we follow the conventional test split, which considers every 8th image as a test image, and compute the SSIM score between the rendered image and ground truth.
Second, we evaluate true novel view renderings that are rendered from a novel trajectory\footnote{Novel trajectories are generated by interpolating training poses with a B-spline function (degree of 10), creating 20 novel poses per scene.} with \ourscore without ground truth.
\cref{tab:exp:novel_trajectory_correlation} indicate a close correlation between \ourscore evaluations of novel views and traditional SSIM scores. Additionally, the rankings of rendering quality for these scenes, determined using both SSIM and \ourscore, are also closely aligned.

\input{figures/03_attn}
\input{figures/02_ablation}
\input{tables/ablation_ref_on_off}

\subsection{Visualising Attention Weights}
\label{sec:exp:vis_attn}
To delve deeper into our cross-reference method, \cref{fig:exp:attn} visualises the attention weights in the cross-attention layer for the central patch of a query image. 
This illustration confirms that the cross-attention mechanism effectively focuses on similar content from the cross-reference set, thereby providing insight into the results of the ablation study in \cref{sec:exp:ablation}.

\subsection{Ablation Study: Enable and Disable Reference Views}
\label{sec:exp:ablation}
This experiment demonstrates that our cross-reference module effectively uses the cross-reference set for quality prediction. 
When provided with reference images, the module offers detailed and accurate evaluations, as shown in \cref{fig:ablation}, in contrast to the high scores predicted across almost all regions when reference images are disabled.
Note that, in this context, we disable reference images by setting all pixels in reference images to zero.
Quantitative support for these findings is presented in \cref{tab:ablation_ref_on_off}.

\subsection{Limitations and Future Work}
\label{sec:exp:limitation_and_future}
\input{figures/05_aria}
We outline two future research directions: 
First, enhancing the sharpness of our score maps to match the clarity of full-reference SSIM, possibly by integrating pixel-level positional encoding or super-resolution methods to mitigate the blurring from patch-wise encoding of ViT models.
Second, tackling the issue with unconventional images, such as those from fish-eye lenses that lead to inaccurate predictions, as illustrated in \cref{fig:exp:aria}.

%% file: figures/04_main_results.tex
\begin{figure}[t]
    \centering
    \includegraphics[width=1.0\columnwidth]{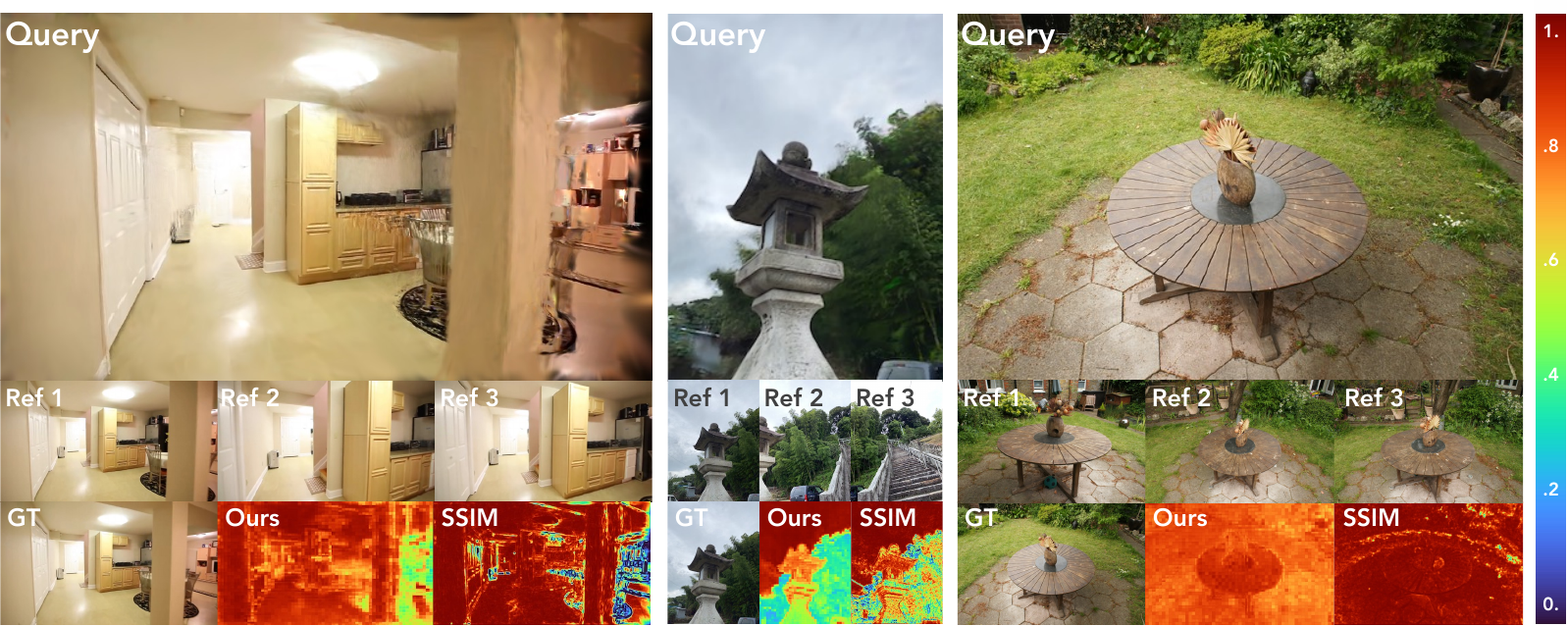}
    \caption{\textbf{Qualitative results of \ourscore and SSIM on various datasets.} We present examples for test results on each dataset (from left to right: RE10K, MFR, Mip360). We show our score maps have a strong correlation with SSIM score, demonstrating the generalisation capability of our approach across diverse datasets.
    Score colour coding: \textcolor{red}{red} represents the highest score, followed by \textcolor{orange}{orange}, \textcolor{green}{green}, and \textcolor{blue}{blue}, indicating decreasing scores respectively.
    }
    \label{fig:exp:main_results}
\end{figure}

%% file: tables/main_correlation.tex
\begin{table}[t]
\centering
\caption{
    \textbf{Correlation between various metrics and SSIM on various datasets.}
    \textbf{FR}: full-reference.
    \textbf{NR}: no-reference.
    \textbf{CR}: cross-reference.
    We show \ourscore is highly correlated with SSIM score on various datasets, while only being trained with the MFR dataset. 
}
\begin{tabular}{llcclccclc}
\toprule
\multirow{2}{*}{Datasets} &  & \multicolumn{2}{c}{FR} &  & \multicolumn{3}{c}{NR}  &  & CR            \\ \cline{3-4} \cline{6-8} \cline{10-10} 
                          &  & SSIM$\uparrow$      & PSNR$\uparrow$      &  & BRISQUE$\downarrow$ & NIQE$\downarrow$ & PIQE$\downarrow$ &  & Ours$\uparrow$          \\ \hline
RE10K                     &  & 1.00       & 0.92      &  & 0.46    & 0.32  & 0.27  &  & \textbf{0.99} \\
Mip360                    &  & 1.00       & 0.91      &  & 0.19    & 0.61  & 0.69  &  & \textbf{0.95} \\
MFR                       &  & 1.00       & 0.92      &  & 0.23    & -0.30 & -0.11 &  & \textbf{0.83} \\ \bottomrule
\end{tabular}
\label{tab:main_correlation}
\end{table}

%% file: tables/ibrnet_vs_pixelnerf.tex

\begin{table}[t]
\centering
\caption{
    \textbf{Evaluating Few-shot NeRFs with Various Metrics.} 
    We show that when comparing two few-shot NeRF models IBRNet~\cite{wang2021ibrnet} and PixelNeRF~\cite{yu2021pixelnerf}, \ourscore is consistent with full-reference metrics such as SSIM and PSNR. In this case, all metrics shows that IBRNet performs better than PixelNeRF on MFR dataset.
    \label{tab:exp:ibrnet_vs_pixelnerf}
}
\begin{tabular}{lccccccccc}
\toprule
\multirow{2}{*}{NVS} & & \multicolumn{2}{c}{FR}            &           & \multicolumn{3}{c}{NR}                                       &           & CR              \\ \cline{3-4} \cline{6-8} \cline{10-10} 
                  & & SSIM$\uparrow$ & PSNR$\uparrow$ &           & BRISQUE$\downarrow$ & NIQE$\downarrow$ & PIQE$\downarrow$ &           & Ours$\uparrow$ \\ \midrule
PixelNeRF         & & 0.26            & 9.17            &           & 35.46                & 5.44              & 35.96             &           & 0.40            \\
IBRNet   & & \textbf{0.44}   & \textbf{18.51}  & \textbf{} & \textbf{23.47}       & \textbf{2.68}     & \textbf{23.35}    & \textbf{} & \textbf{0.71}   \\ \bottomrule
\end{tabular}
\end{table}

%% file: figures/06_novel_traj.tex
\begin{figure}[t]
    \centering
    \includegraphics[width=1.0\columnwidth]{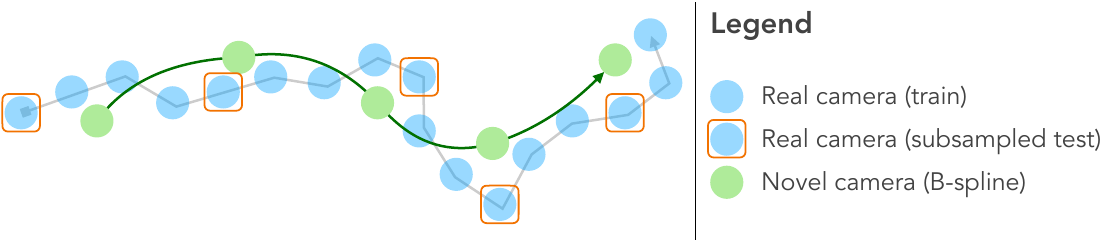}
    \caption{
    \textbf{Illustration of two IQA approaches in NVS: 1) with subsampled test views and 2) with true novel views.} 
    The first approach relies on full-reference metrics that requires ground truth images, precluding test views in training (\textcolor{CornflowerBlue}{blue} circles enclosed in \textcolor{orange}{orange} boxes).
    In contrast, our cross-reference approach bypasses the need for ground truth views, allowing NVS evaluation from true novel views (\textcolor{ForestGreen}{green} circles) and enabling NVS modelling to utilise the entire captured image set.
    }
    \label{fig:exp:novel_traj_setup}
\end{figure}

%% file: tables/novel_trajectory.tex
\begin{table}[t]
\centering
\caption{    
    \textbf{IQA on Images Renderings From Novel Trajectories.}
    We evaluate each sequence in two ways: 
        1) computing SSIM on images rendered from the standard subsampled test split with ground truth images, and 
        2) computing \ourscore on images rendered from a novel trajectory, with a cross-reference set randomly sampled from training images.
    We show that our method can evaluate the quality of Gaussian-Splatting from a novel trajectory without requiring aligned ground truth images. 
    Our cross-reference style score is highly correlated with the full-reference SSIM score, and ranking video quality using \ourscore is similar to ranking with SSIM.
    \textbf{Top}: SSIM and \ourscore. Higher is better.
    \textbf{Bottom}: quality ranking using SSIM and \ourscore respectively. Lower is better.
    `Corr' denotes Pearson correlation for scores and Spearman's rank correlation for rankings.
}
\label{tab:exp:novel_trajectory_correlation}
\resizebox{\textwidth}{!}{%
\begin{tabular}{@{}lllcccccccccccccccccccccccccccccc@{}}
\toprule
                                   &  & Scene &  & 426  &  & 34   &  & 10   &  & 135  &  & 238  &  & 284  &  & 103  &  & 441  &  & 345  &  & 311  &  & 175  &  & 244  &  & 82   &  & 4    &  & \textbf{Corr}                  \\ \midrule
\multirow{2}{*}{Score $\uparrow$}  &  & SSIM  &  & 0.74 &  & 0.66 &  & 0.64 &  & 0.64 &  & 0.61 &  & 0.61 &  & 0.59 &  & 0.58 &  & 0.56 &  & 0.55 &  & 0.51 &  & 0.50 &  & 0.44 &  & 0.40 &  & \multirow{2}{*}{\textbf{0.84}} \\
                                   &  & Ours  &  & 0.80 &  & 0.78 &  & 0.77 &  & 0.78 &  & 0.66 &  & 0.61 &  & 0.73 &  & 0.75 &  & 0.73 &  & 0.72 &  & 0.62 &  & 0.58 &  & 0.55 &  & 0.53 &  &                                \\ \midrule
\multirow{2}{*}{Rank $\downarrow$} &  & SSIM  &  & 0    &  & 1    &  & 2    &  & 3    &  & 4    &  & 5    &  & 6    &  & 7    &  & 8    &  & 9    &  & 10   &  & 11   &  & 12   &  & 13   &  & \multirow{2}{*}{\textbf{0.85}} \\
                                   &  & Ours  &  & 0    &  & 2    &  & 3    &  & 1    &  & 8    &  & 10   &  & 6    &  & 4    &  & 5    &  & 7    &  & 9    &  & 11   &  & 12   &  & 13   &  &                                \\ \bottomrule
\end{tabular}%
}
\end{table}

%% file: figures/03_attn.tex
\begin{figure}[t]
    \centering
    \begin{subfigure}[b]{0.48\columnwidth}
        \includegraphics[width=\columnwidth]{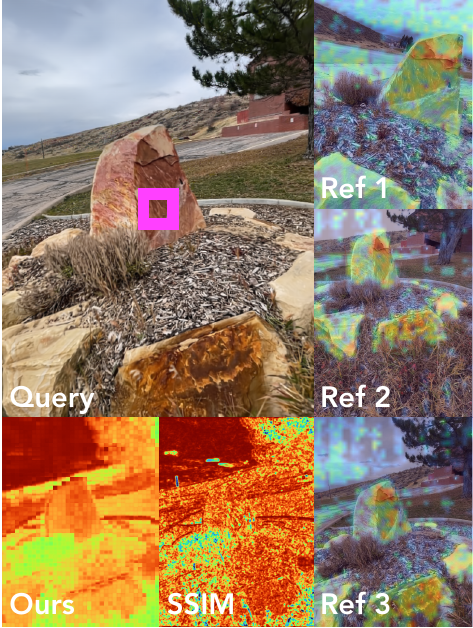}
    \end{subfigure}
    \begin{subfigure}[b]{0.48\columnwidth}
        \includegraphics[width=\columnwidth]{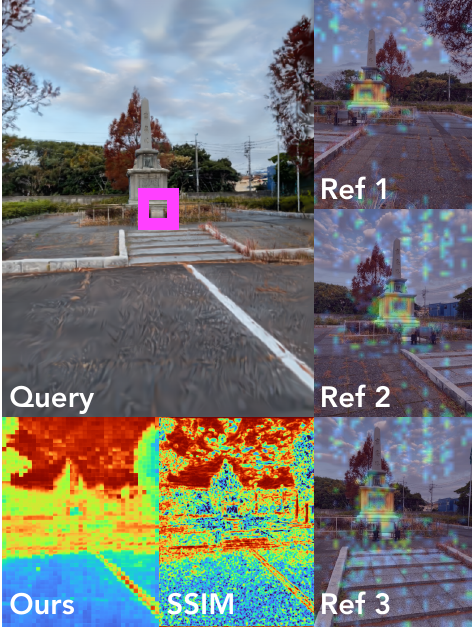}
    \end{subfigure}
    \caption{
        \textbf{Visualisation of attention weights from the cross-reference module $\Phi_{\text{cross}}$.}
        \textbf{Top left}: a query image with a region of interest (centre of image) highlighted with a \textcolor{magenta}{magenta} box.
        \textbf{Right column}: We show 3 reference images from our cross-reference set with attention maps overlaid. The attention maps illustrate the attention that is paid to predicting image quality at the query region.
        \textcolor{red}{Red} and \textcolor{blue}{blue} denote high and low attention weights respectively. Note that we use $N_\text{ref}=5$ but only 3 is shown due to space constraint.
        \textbf{Bottom}: Predicted \ourscore map and SSIM map. \textcolor{red}{Red} and \textcolor{blue}{blue} denote high and low quality image regions respectively.
    }
    \label{fig:exp:attn}
\end{figure}

%% file: figures/02_ablation.tex
\begin{figure}[h!]
    \centering
    \includegraphics[width=1.0\columnwidth]{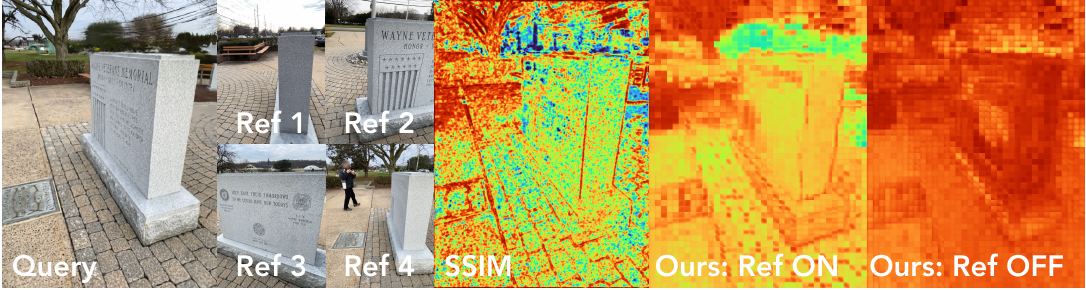}
    \caption{
        \textbf{Ablation study: reference set enabled (on) and disabled (off).}
        We show that with reference images enabled, the score map predicted by our method contains more details. When the reference images are disabled, the model tends to assign everything a high score. This is also evidenced by quantitative results in \cref{tab:ablation_ref_on_off}.
    }
    \label{fig:ablation}
\end{figure}

%% file: tables/ablation_ref_on_off.tex
\begin{table}[t]
\centering
\caption{    
    \textbf{Ablation study: reference set enabled (\cmark) and disabled (\xmark).}
    Our method performs closer to SSIM when reference images are provided. Note that when reference images are disabled, the predicted scores still show a certain level of correlation, as certain noise patterns can be identified from local image statistics. In this case our method degrades to a \textit{no-reference}-style image evaluation. 
}
\resizebox{\textwidth}{!}{%
\begin{tabular}{lcccccccccccccccccccccccccccccccc}
\toprule
Scene &  & 4    &  & 10   &  & 34   &  & 82   &  & 103  &  & 135  &  & 175  &  & 238  &  & 244  &  & 284  &  & 311  &  & 345  &  & 426  &  & 441  &  & Avg &  & \textbf{Corr} \\ \hline
SSIM  &  & 0.40 &  & 0.64 &  & 0.66 &  & 0.44 &  & 0.59 &  & 0.64 &  & 0.51 &  & 0.61 &  & 0.50 &  & 0.61 &  & 0.55 &  & 0.56 &  & 0.74 &  & 0.58 &  & 0.57 &  & \textbf{1.00} \\ \hline
Ours \cmark  &  & 0.46 &  & 0.72 &  & 0.72 &  & 0.48 &  & 0.64 &  & 0.75 &  & 0.56 &  & 0.61 &  & 0.66 &  & 0.58 &  & 0.65 &  & 0.72 &  & 0.82 &  & 0.71 &  & 0.65 &  & \textbf{0.83} \\
Ours \xmark  &  & 0.71 &  & 0.81 &  & 0.83 &  & 0.73 &  & 0.80 &  & 0.84 &  & 0.80 &  & 0.79 &  & 0.86 &  & 0.80 &  & 0.74 &  & 0.86 &  & 0.89 &  & 0.85 &  & 0.81 &  & \textbf{0.68} \\ \bottomrule
\end{tabular}%
}
\label{tab:ablation_ref_on_off}
\end{table}

%% file: figures/05_aria.tex
\begin{wrapfigure}{r}{0.29\columnwidth}
    \vspace{-1.5cm}
    \centering
    \includegraphics[width=0.29\columnwidth]{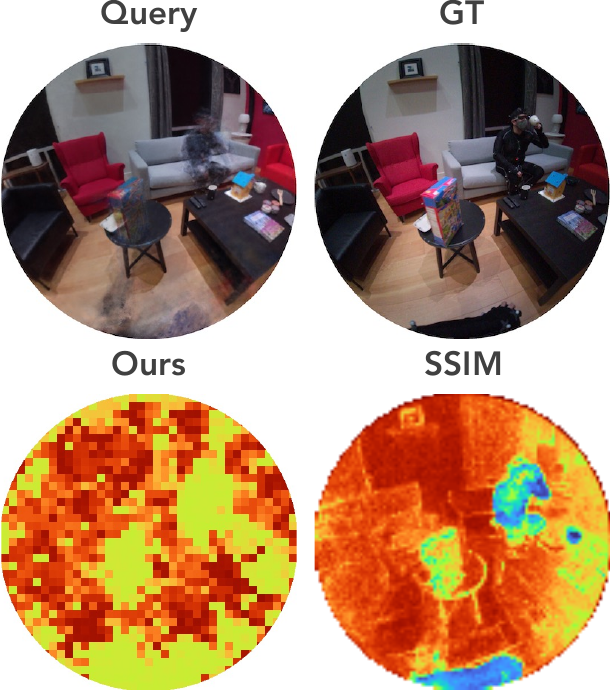}
    \caption{Evaluating a fish-eye-style query image.}
    \vspace{-1.5cm}
    \label{fig:exp:aria}
\end{wrapfigure}


%% file: 05_conclusion.tex
\section{Conclusion}
\label{sec:conclusion}

In summary, we introduce a novel Cross-Reference Image Quality Assessment (CR-IQA) scheme, filling a critical gap in existing IQA schemes. 
By leveraging a neural network with cross-attention mechanisms and a unique NVS-enabled data collection pipeline, we demonstrate the feasibility of accurately evaluating the quality of an image by comparing it with other views of the same scene.
Our experimental results indicate that our predictions closely align with ground-truth-dependent metrics.

%% file: 06_ack.tex
\section*{Acknowledgement}
This research is supported by an ARIA research gift grant from Meta Reality Lab.
We gratefully thank Shangzhe Wu, Tengda Han, Zihang Lai for insightful discussions, and Michael Hobley for proofreading.

%% file: 08_supp.tex
\clearpage
{
\begin{center}
\Large
\textbf{\texttt{CrossScore}: Towards Multi-View Image Evaluation and Scoring \\(Supplementary)}
\end{center}
}

{
\begin{center}
Zirui Wang \quad
Wenjing Bian \quad
Victor Adrian Prisacariu\\[5pt]
University of Oxford
\end{center}
}

\appendix

\section{Additional Practical Details}
At test time, evaluating images at resolution 518$\times$690 with batch size 16 takes 20GB GPU memory with an average time consumption of 69ms/img on an NVIDIA RTX 4090 GPU.
Our model consists of 26M parameters, of which 23M are from the pre-trained DINOv2-small backbone $\Phi_{\text{enc}}$, and 3M are from our cross attention module $\Phi_{\text{cross}}$ and our score regression head $\Phi_{\text{dec}}$.

\section{Additional Quantitative Details for \Cref{tab:main_correlation}}
We offer detailed results for \cref{tab:main_correlation} in \Cref{tab:supp:corr_main_detail_MFR,tab:supp:corr_main_detail_mip360,tab:supp:corr_main_detail_RE10K}.
For no-reference baselines, assessments are conducted using Matlab with their default settings and feature models.

\begin{table}[h]
\centering
\caption{Correlation between various metrics and SSIM on the Map-Free Relocalisation (MFR) dataset.}
\begin{tabular}{llcclccclc}
\multicolumn{10}{c}{\textbf{MFR}}                                                         \\ \hline
\multirow{2}{*}{Scene} &  & \multicolumn{2}{c}{FR} &  & \multicolumn{3}{c}{NR}  &  & CR   \\ \cline{3-4} \cline{6-8} \cline{10-10} 
                       &  & SSIM $\uparrow$     & PSNR $\uparrow$       &  & BRISQUE $\downarrow$ & NIQE $\downarrow$ & PIQE $\downarrow$ &  & Ours $\uparrow$ \\ \hline
s00004                 &  & 0.40      & 15.88      &  & 19.40   & 3.00  & 34.10 &  & 0.46 \\
s00010                 &  & 0.64      & 19.16      &  & 20.82   & 3.21  & 28.39 &  & 0.72 \\
s00034                 &  & 0.66      & 21.91      &  & 25.23   & 2.66  & 29.74 &  & 0.72 \\
s00082                 &  & 0.44      & 16.35      &  & 23.47   & 2.68  & 34.54 &  & 0.48 \\
s00103                 &  & 0.59      & 16.43      &  & 30.50   & 3.16  & 47.39 &  & 0.64 \\
s00135                 &  & 0.64      & 20.12      &  & 20.90   & 2.88  & 42.21 &  & 0.75 \\
s00175                 &  & 0.51      & 17.32      &  & 24.83   & 3.24  & 31.79 &  & 0.56 \\
s00238                 &  & 0.61      & 16.74      &  & 27.15   & 2.74  & 34.17 &  & 0.61 \\
s00244                 &  & 0.50      & 18.03      &  & 25.57   & 3.24  & 42.06 &  & 0.66 \\
s00284                 &  & 0.61      & 19.46      &  & 25.79   & 2.78  & 30.60 &  & 0.58 \\
s00311                 &  & 0.55      & 17.82      &  & 24.08   & 3.21  & 26.75 &  & 0.65 \\
s00345                 &  & 0.56      & 18.82      &  & 19.71   & 2.83  & 41.05 &  & 0.72 \\
s00426                 &  & 0.74      & 22.10      &  & 24.41   & 2.66  & 32.16 &  & 0.82 \\
s00441                 &  & 0.58      & 20.36      &  & 26.03   & 2.51  & 39.36 &  & 0.71 \\ \hline
Correlation            &  & 1.00      & 0.78       &  & 0.23    & -0.30 & -0.11 &  & 0.83 \\ \bottomrule
\end{tabular}
\label{tab:supp:corr_main_detail_MFR}
\end{table}

\begin{table}[t]
\centering
\caption{Correlation between various metrics and SSIM on the Mip360 dataset.}
\begin{tabular}{llcclccclc}
\multicolumn{10}{c}{\textbf{Mip360}}                                                     \\ \hline
\multirow{2}{*}{Scene} &  & \multicolumn{2}{c}{FR} &  & \multicolumn{3}{c}{NR} &  & CR   \\ \cline{3-4} \cline{6-8} \cline{10-10} 
                       &  & SSIM $\uparrow$     & PSNR $\uparrow$      &  & BRISQUE $\downarrow$ & NIQE $\downarrow$ & PIQE $\downarrow$ &  & Ours $\uparrow$ \\ \hline
bicycle                &  & 0.85      & 26.66      &  & 22.30   & 2.67 & 33.41 &  & 0.82 \\
bonsai                 &  & 0.95      & 32.48      &  & 27.08   & 3.46 & 52.90 &  & 0.89 \\
counter                &  & 0.92      & 29.82      &  & 25.61   & 2.78 & 50.57 &  & 0.87 \\
flowers                &  & 0.72      & 25.31      &  & 26.53   & 2.57 & 31.46 &  & 0.64 \\
garden                 &  & 0.92      & 31.19      &  & 13.18   & 2.37 & 31.38 &  & 0.87 \\
kitchen                &  & 0.95      & 32.48      &  & 30.73   & 3.03 & 43.82 &  & 0.85 \\
room                   &  & 0.94      & 33.42      &  & 33.43   & 2.93 & 53.95 &  & 0.91 \\
stump                  &  & 0.83      & 30.43      &  & 22.52   & 2.97 & 21.35 &  & 0.81 \\
treehill               &  & 0.74      & 25.25      &  & 23.58   & 2.30 & 31.22 &  & 0.73 \\ \hline
Correlation            &  & 1.00      & 0.91       &  & 0.19    & 0.61 & 0.69  &  & 0.95 \\ \bottomrule
\end{tabular}
\label{tab:supp:corr_main_detail_mip360}
\end{table}

\begin{table}[h!]
\centering
\caption{Correlation between various metrics and SSIM on the RealEstate10K (RE10K) dataset.}
\begin{tabular}{llcclccclc}
\multicolumn{10}{c}{\textbf{RealEstate10K}}                                              \\ \hline
\multirow{2}{*}{Scene} &  & \multicolumn{2}{c}{FR} &  & \multicolumn{3}{c}{NR} &  & CR   \\ \cline{3-4} \cline{6-8} \cline{10-10} 
                       &  & SSIM $\uparrow$     & PSNR $\uparrow$      &  & BRISQUE $\downarrow$ & NIQE $\downarrow$ & PIQE $\downarrow$ &  & Ours $\uparrow$ \\ \hline
00407b3f1bad1493       &  & 0.90      & 26.06      &  & 44.33   & 3.70 & 69.66 &  & 0.91 \\
004ed278c2b168f1       &  & 0.73      & 20.13      &  & 53.48   & 4.39 & 54.82 &  & 0.77 \\
0065a058603dfca4       &  & 0.88      & 22.98      &  & 49.01   & 4.18 & 75.67 &  & 0.90 \\
00703cbf7531ef11       &  & 0.56      & 17.74      &  & 30.67   & 2.57 & 42.89 &  & 0.67 \\
00761c6dcec91853       &  & 0.95      & 31.34      &  & 44.70   & 3.83 & 62.81 &  & 0.93 \\
007ac6cef80a692c       &  & 0.90      & 22.76      &  & 33.68   & 3.25 & 70.71 &  & 0.91 \\
0081cfd790d7ad74       &  & 0.02      & 10.45      &  & NaN     & NaN  & NaN   &  & 0.23 \\
009664cb1b8d351a       &  & 0.74      & 17.00      &  & 52.77   & 4.39 & 82.10 &  & 0.74 \\
00a50bfbce75d465       &  & 0.86      & 23.86      &  & 38.82   & 3.22 & 65.02 &  & 0.88 \\
00a9f110ad222aa4       &  & 0.81      & 22.34      &  & 32.61   & 2.25 & 49.61 &  & 0.82 \\
00b52b21e0d54a42       &  & 0.89      & 22.64      &  & 43.64   & 3.70 & 72.21 &  & 0.90 \\
00b9a7963f9bd9c6       &  & 0.37      & 14.60      &  & 34.06   & 3.30 & 67.52 &  & 0.38 \\
00c8250efd605554       &  & 0.15      & 8.73       &  & 32.17   & 2.98 & 60.74 &  & 0.19 \\ \hline
Correlation            &  & 1.00      & 0.92       &  & 0.46    & 0.32 & 0.27  &  & 0.99 \\ \bottomrule
\end{tabular}
\label{tab:supp:corr_main_detail_RE10K}
\end{table}

\section{Additional Qualitative Results}
We invite readers to check out a video with additional  qualitative results on our project page: \url{https://crossscore.active.vision}.

\clearpage
\section{Discussion: Relationships with Visual Place Recognition (VPR) Systems}
One straightforward way to evaluate an image with a full-reference metric such as SSIM without aligned ground truth is to utilise a nearby frame, for example, a temporal neighbour, or a visually similar frame obtained from image retrieval or visual place recognition (VPR) systems.
\cref{fig:supp:vpr,tab:supp:vpr} demonstrate that SSIM scores computed using misaligned images (nearest frames) are significantly different from GT SSIM scores, whereas our multi-view-based scores are similar to GT SSIM scores.
\input{figures/07_vpr}
\input{tables/vpr}

\section{Social Impact}
Our cross-reference image quality assessment method has limited negative social impact. It enhances image evaluations for applications like novel view synthesis without using human data, thus avoiding privacy issues. 
Our method does not facilitate harmful activities and focuses on technical improvements. With low misuse potential and significant benefits for fields like computer graphics and virtual reality, this advancement positively impacts technological and creative industries without significant ethical concerns.

%% file: figures/07_vpr.tex
\begin{figure}[h]
    \centering
    \includegraphics[width=1.0\columnwidth]{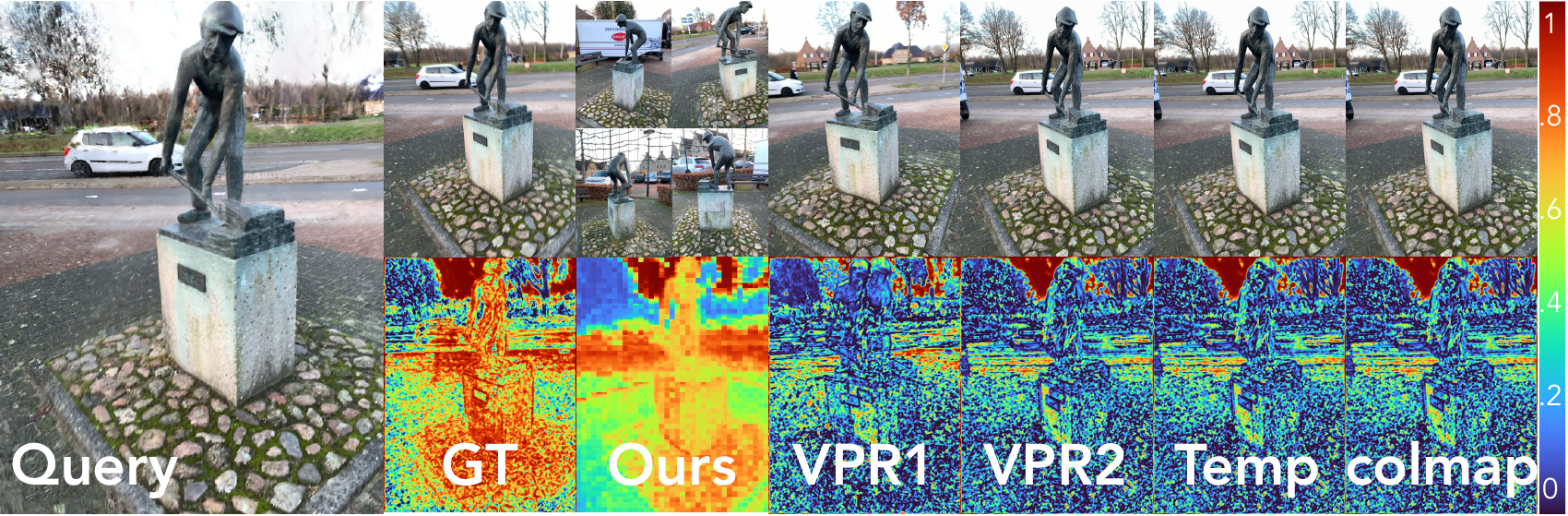}
    \caption{
    \textbf{Our method vs. computing SSIM between a query image and a nearest frame}. The nearest frame is selected through various strategies: 
    \textbf{VPR1}: SALAD~\cite{izquierdo2024optimal}, 
    \textbf{VPR2}: CricaVPR~\cite{lu2024cricavpr}, 
    \textbf{Temp}: temporal nearest frame, and
    \textbf{colmap}: vocabulary-tree-based image retrieval module in COLMAP~\cite{schonberger2016structure}.
    We show that the SSIM scores computed using misaligned images (nearest frames) are significantly different from GT SSIM scores, whereas our multi-view-based scores are similar to GT SSIM scores.
    }
    \label{fig:supp:vpr}
\end{figure}

%% file: tables/vpr.tex
\begin{table}[h]
\centering
\caption{
    \textbf{Correlation between GT SSIM, our score, and SSIM scores computed using various nearest frames.}
    Each nearest frame is selected through the following strategies:  
    \textbf{VPR1}: SALAD~\cite{izquierdo2024optimal}, 
    \textbf{VPR2}: CricaVPR~\cite{lu2024cricavpr}, 
    \textbf{Temp}: temporal nearest frame, and
    \textbf{COLMAP}: vocabulary-tree-based image retrieval module in COLMAP~\cite{schonberger2016structure}.
    We show that the SSIM scores computed using misaligned images (nearest frames) are significantly different from GT SSIM scores, whereas our multi-view-based scores are similar to GT SSIM scores.
    }
\label{tab:supp:vpr}
\begin{tabular}{cccccccccccc}
\toprule
& GT & & Ours   & & VPR1~\cite{izquierdo2024optimal} & & VPR2~\cite{lu2024cricavpr} & & Temp & & COLMAP~\cite{schonberger2016structure} \\ \midrule
Corr & 1.0 & & \textbf{0.83} & & 0.37  & & 0.38     & & 0.37     & & 0.39   \\ \bottomrule
\end{tabular}%
\end{table}

%% file: 07_reference.tex
%
%
\bibliographystyle{splncs04}
\bibliography{ref}